\documentclass[runningheads]{llncs}

\usepackage[T1]{fontenc}
\usepackage{graphicx}
\usepackage{booktabs}

\usepackage{amsfonts,amsmath}
\usepackage{algorithm,algpseudocode}
\usepackage{mathrsfs}
\usepackage{nicefrac}
\usepackage{subcaption}
\usepackage{multirow}
\usepackage{xcolor}
\usepackage{comment}

\usepackage{makecell}
\usepackage{hyperref}

\usepackage[misc]{ifsym}

\usepackage{mwe}

\usepackage[misc]{ifsym}

\begin{document}

\title{Computing Gram Matrix for SMILES Strings using RDKFingerprint and Sinkhorn-Knopp Algorithm}



\author{}
\institute{}

\author{Sarwan Ali\inst{1} \and
Haris Mansoor\inst{2} \and
Prakash Chourasia \inst{1}
Imdad Ullah Khan \inst{2}
\and Murray Pattersn \inst{1}
}
\authorrunning{S. Ali et al.}
%
\institute{Georgia State University, Atlanta GA 30303, USA \and
Lahore University of Management Sciences, Lahore Punjab 54792, Pakistan
\email{sali85@student.gsu.edu, 16060061@lums.edu.pk, pchourasia1@student.gsu.edu, imdad.khan@lums.edu.pk, mpatterson30@gsu.edu}}

\maketitle              
\begin{abstract}
In molecular structure data, SMILES (Simplified Molecular Input Line Entry System) strings are used to analyze molecular structure design. Numerical feature representation of SMILES strings is a challenging task. This work proposes a kernel-based approach for encoding and analyzing molecular structures from  SMILES strings. The proposed approach involves computing a kernel matrix using the Sinkhorn-Knopp algorithm while using kernel principal component analysis (PCA) for dimensionality reduction. The resulting low-dimensional embeddings are then used for classification and regression analysis.
The kernel matrix is computed by converting the SMILES strings into molecular structures using the Morgan Fingerprint, which computes a fingerprint for each molecule. The distance matrix is computed using the pairwise kernels function.
The Sinkhorn-Knopp algorithm is used to compute the final kernel matrix that satisfies the constraints of a probability distribution. This is achieved by iteratively adjusting the kernel matrix until the marginal distributions of the rows and columns match the desired marginal distributions. We provided a comprehensive empirical analysis of the proposed kernel method to evaluate its goodness with greater depth.
The suggested method is assessed for drug subcategory prediction (classification task) and solubility AlogPS ``Aqueous solubility and Octanol/Water partition coefficient"  (regression task) using the benchmark SMILES string dataset. The outcomes show the proposed method outperforms several baseline methods in terms of supervised analysis and has potential uses in molecular design and drug discovery. Overall, the suggested method is a promising avenue for kernel methods-based molecular structure analysis and design.
\end{abstract}

\keywords{Kernel Matrix, Classification, Sinkhorn Knopp Algorithm}

\section{Introduction}
Numerous machine learning applications, such as domain adaptation and generalization, heavily rely on numerical representations of the data~\cite{glorot2011domain}. In the field of drug discovery and molecular design, the analysis of molecular structure is a fundamental task~\cite{sellwood2018artificial}. Simplified Molecular Input Line Entry System (SMILES) strings have emerged as a popular choice for representing molecular structure data~\cite{schwaller2022machine}, primarily due to their simplicity and ease of use.
However, modeling and analyzing molecular structures represented as SMILES strings pose several challenges~\cite{krenn2020self}. These include dealing with the high dimensionality of the data and the complex non-linear relationships between the structures. Moreover, converting SMILES strings into machine-readable numerical representations is a challenging task that requires sophisticated techniques.


The analysis of SMILES strings has become increasingly important in the field of drug discovery and cheminformatics~\cite{chen2018cheminformatics}. SMILES strings are a compact representation of a molecule's structure and have become a popular choice for encoding molecular information in machine learning models~\cite{wigh2022review}. These models are used for a range of tasks, including drug solubility~\cite{francoeur2021soltrannet} and subtype prediction. However, there is still a need for further investigation into the effectiveness of different types of embeddings, classification, and regression models for SMILES string analysis. The aim of this research project is twofold: (i) to address this gap in knowledge by evaluating the performance of various embedding methods and machine learning models for classification and regression tasks using SMILES strings as input, (ii) to propose a method for SMILES string analysis. The results of this study could have significant implications for drug discovery research and help to identify the most effective methods for predicting molecular properties.

In this paper, we propose a kernel-based approach for encoding and analyzing molecular structures represented as SMILES strings. Performing two tasks (i) the drug subcategories prediction (classification task) and (ii) the solubility AlogPS (Aqueous solubility and Octanol/Water partition coefficient) prediction (regression task). 
The proposed approach involves computing a kernel matrix using the Sinkhorn-Knopp algorithm~\cite{knight2008sinkhorn} and using kernel principal component analysis (PCA)~\cite{fu2011combination} to reduce the dimensionality of the molecular structures. The resulting low-dimensional embeddings are then used for classification and regression analysis.
The proposed approach starts with converting the SMILES strings into molecular structures using the RDKit library, later we convert them into feature vectors using the RDKFingerprint function, which computes a fingerprint for each molecule. The pairwise kernels function is then used to compute the distance matrix.
We present a novel method for creating a kernel matrix by utilizing an optimal transport matrix~\cite{mialon2020trainable}. The Sinkhorn-Knopp algorithm is used to compute the final kernel matrix that satisfies the constraints of a probability distribution. Achieved iteratively by adjusting the kernel matrix until the marginal distributions of the rows and columns of the matrix match the desired marginal distributions.
Kernel PCA reduces the dimensionality of the molecular structures by projecting the kernel matrix onto a lower-dimensional (LD) space. The resulting LD embeddings capture the intrinsic properties of the molecular structures, making them suitable for classification and regression analysis.

The proposed approach has several potential applications in drug discovery and molecular design. It can be used to analyze large datasets of molecular structures and to identify compounds with desirable properties. It can also be used to design new molecules with desired properties by generating low-dimensional embeddings and using them to search for molecules with similar properties. 
In summary, our contributions to this paper are the following:
\begin{enumerate}
    \item We propose a novel kernel function-based approach for SMILES string analysis using classification and regression. Our method is based on the idea of first converting SMILES strings into molecular graphs, computing features, and using optimal transport matrix, we generate the final kernel matrix.
    \item Using empirical analysis, we show that our kernel-based approach achieves higher classification accuracy and comparable regression performance on benchmark SMILES string dataset.
\end{enumerate}

\section{Related Work}\label{sec_related_work}
Molecular fingerprints are binary vectors used to encode a molecule's structural information~\cite{wigh2022review}. They capture substructures and are useful for predicting molecular properties. Recent studies have explored using fingerprints and embeddings to predict drug solubility~\cite{nakajima2021machine}. Random forest regression and support vector regression outperformed other models in predicting solubility~\cite{chen2018rise}. Graph convolutional neural networks achieved an R-squared value of 0.75 on a dataset of 1144 compounds~\cite{rupp2012fast}. More research is needed to evaluate the effectiveness of different embeddings, classification, and regression models for solubility and drug subtype prediction.
Kernel-based approaches have been widely used in ML and data analysis to capture complex relationships between data points~\cite{qiu2016survey}. 
One popular kernel method for molecular data analysis is the kernel ridge regression (KRR) algorithm~\cite{fabregat2022metric}. KRR is a supervised learning algorithm that uses a kernel matrix to represent the pairwise similarities between molecular structures. The KRR algorithm has been applied to several molecular property prediction tasks, including drug solubility, and toxicity prediction.
Another popular kernel method for molecular data analysis is the support vector machine (SVM) algorithm~\cite{thomas2017multi}. SVM is a widely used supervised learning algorithm that uses a kernel matrix to represent the pairwise similarities between data points. Several studies have applied SVM to molecular activity classification tasks, including predicting protein-ligand binding affinity~\cite{shen2020machine} and identifying active compounds in high-throughput screening experiments~\cite{li2009novel}.
Recently, kernel principal component analysis (PCA) has emerged as a powerful tool for dimensionality reduction and feature extraction in molecular data analysis~\cite{rensi2017flexible,fu2011combination}. Kernel PCA uses a kernel matrix to project the high-dimensional molecular structures onto a lower-dimensional space, capturing the intrinsic properties of the molecular structures. Several studies use kernel PCA for molecular property prediction~\cite{fu2011combination,rensi2017shallow} and activity classification tasks, and have shown that it can significantly improve the performance of existing methods.
While these methods have shown promising results, they have several limitations (i.e. do not capture the intrinsic properties of the molecular structures and may suffer from overfitting). 

\section{Proposed Approach}\label{sec_proposed_approach}
The proposed approach takes a pair of SMILES strings $S_1$ and $S_2$ as input and computes a kernel value between the molecules represented by the SMILES strings. The kernel value is computed using the following steps described below.

\subsection{Convert the SMILES strings into molecular graphs}
Let $X_1$ and $X_2$ be the sets of molecular graphs corresponding to the SMILES strings in $S_1$ and $S_2$, respectively. For each SMILES string, we use the MolFromSmiles function from the chem library in the RDKit library to obtain the corresponding molecular graph. This gives us the sets of molecular graphs $X_1 = {x_{1,1}, x_{1,2}, \dots, x_{1,n}}$ and $X_2 = {x_{2,1}, x_{2,2}, \dots, x_{2,n}}$.

\subsection{Compute the molecular features}
Next, we convert each molecular graph $x_{i,j}$ into a feature vector using the RDKit fingerprint. The fingerprint is a bit vector of a fixed length representing the presence or absence of certain molecular substructures in the molecule. The RDKit fingerprint is generated using the Morgan algorithm~\cite{nakajima2021machine}, which is a circular fingerprinting method that considers the neighborhood of each atom in the molecule up to a certain radius. The resulting fingerprint can be used to compare the structural similarity of different molecules, among other applications in cheminformatics.
This fingerprint gives us the sets of feature vectors $X_1^{feat} = {f_{1,1}, f_{1,2}, \dots, f_{1,n}}$ and $X_2^{feat} = {f_{2,1}, f_{2,2}, \dots, f_{2,n}}$, where $n = 2048$, which is defined by Morgan Fingerprint.

\subsection{Compute the pairwise distance matrix}
We compute the pairwise distance matrix $D$ between the feature vectors using the Gaussian kernel of width $\sigma$:
\begin{equation}
\label{eq:1}
    D_{i,j} = exp ( - \frac{\vert \vert X_i^{feat} - X_j^{feat} \vert \vert^2}{2 \sigma^2})
\end{equation}
where $D_{i,j}$ is the $i,j^{th}$ entry of the distance matrix $D$. Iterating over all $i$ and $j$ will give us a $N \times N$ matrix $D$, where $N$ is the number of SMILES strings.

\subsection{Compute the kernel matrix}
We use the Sinkhorn-Knopp algorithm~\cite{knight2008sinkhorn} to compute the optimal transport matrix between the two sets of molecules. Let $K_{i,j}$ be the entry in the kernel matrix $K$ corresponding to molecules $x_{1,i}$ and $x_{2,j}$. We first normalize the pairwise distance matrix $D$ to obtain a joint probability matrix $P$:
\begin{equation} 
\label{eq:2}
    P_{i,j} = \frac{ D_{i,j} / \sigma}{\sum_{k,l} D_{k,l}}
\end{equation}

We then use zero vectors $\overrightarrow{a}, \overrightarrow{b}$ and unit vectors  $\overrightarrow{a_1}, \overrightarrow{b_1}$ and iteratively compute a bipartite graph using the probability matrix. More formally:
\begin{equation} 
\label{eq:3}
BipartiteGraph \gets \frac{-\delta P + (\zeta \times log(\overrightarrow{a_1})) + (\zeta \times log(\overrightarrow{b_1}))}{\zeta}
 \end{equation}
where $\zeta= 1, \delta = 10^{-10}$. Note that the $\xi$ is the tolerance parameter. 
The $\overrightarrow{a_1}$ and $\overrightarrow{b_1}$ are linear sums of the rows and columns of the bipartite graph, respectively, which are then used to update $\overrightarrow{a}$ and $\overrightarrow{b}$. This process is iteratively continued until the convergence criteria are met. The convergence criteria is $max(\vert \overrightarrow{a_{1}} - \overrightarrow{a} \vert) < \xi$ and  $max(\vert \overrightarrow{b_{1}} - \overrightarrow{b} \vert) < \xi$. Note that $\xi$ is a hyperparameter whose value is the following $\xi = 10^{-6}$. 
Finally, we compute the kernel matrix ${K}$ using the probability matrix:  
\begin{equation} 
\label{eq:4}
    K = a' \times P \times b'
\end{equation}

where $a'$ and $b'$ are both diagonal matrix version of $\overrightarrow{a}$ and $\overrightarrow{b}$.

\begin{figure}[h!]
  \centering
        \includegraphics[scale=0.22]{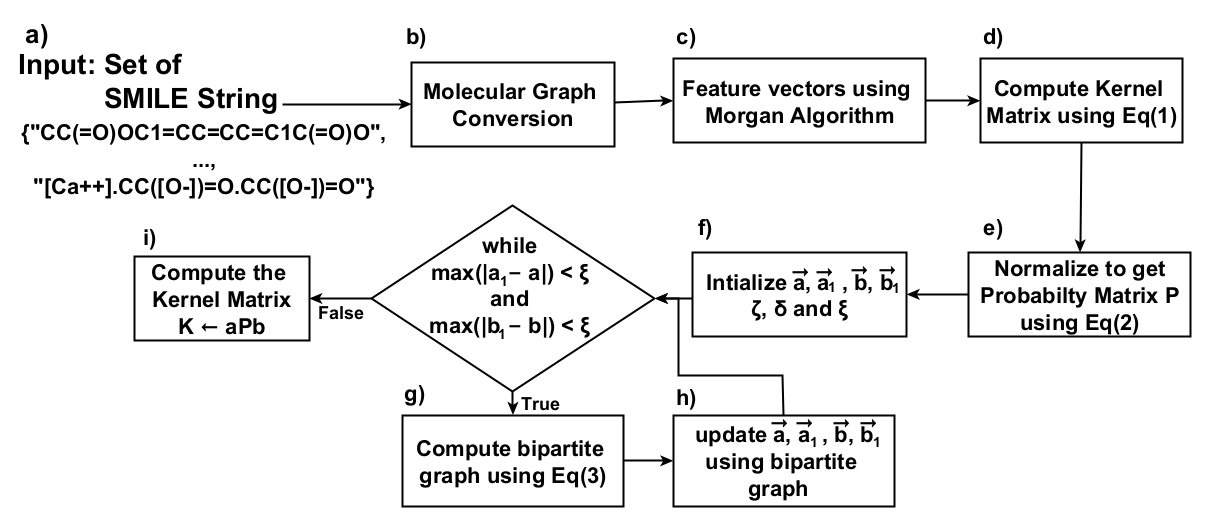}
    \caption{Workflow of the proposed method.}
    \label{Workflow_kernel}
\end{figure}

The overall workflow is given in Figure~\ref{Workflow_kernel}. For a set of input SMILES strings, the first step is to convert them to molecular graphs and compute feature vectors using the Morgan algorithm (RDKFingerprint)~\cite{nakajima2021machine}, Figure~\ref{Workflow_kernel} (b), and (c). 
In the next step, the feature vectors are converted into Gaussian kernel matrix $D$ Figure~\ref{Workflow_kernel} (d), and normalized to transform into a probability matrix $P$, Figure~\ref{Workflow_kernel} (e). Then we we initialize vectors $a, a_{1}$, $b, b_{1}$ and use the
Sinkhorn-Knopp algorithm~\cite{knight2008sinkhorn} to compute the optimal transport matrix $a$ and $b$ iteratively till convergence criteria are met. We use bipartite graph~\ref{eq:3} also shown in  Figure~\ref{Workflow_kernel} (g) and (h) to compute the matrix. After these matrix $a$ and $b$ returned by the Sinkhorn-Knopp algorithm are then multiplied with the probability matrix P to compute the final kernel matrix $K$ (in Eq~\ref{eq:4}) Figure~\ref{Workflow_kernel} (i).

\subsection{Kernel PCA-Based Embeddings}
In practice, a large dataset leads to a large K, and storing K may become a problem. Kernel-PCA can help in this regard to convert K into a low-dimensional subspace. Since our data is highly non-linear, kernel PCA can find the non-linear manifold. Using the kernel, the originally linear operations of PCA are performed in a reproducing kernel Hilbert space. It does so by mapping the data into a higher-dimensional space but then turns out to lie in a lower-dimensional subspace of it. So kernel-PCA increases the dimensionality in order to be able to decrease it. 
Using kernel-PCA, we compute the top principal components from K. These principal components are used as the feature vectors, which can then be used as input for any linear and nonlinear classifiers as well as regression models for supervised analysis.

\section{Experimental Setup}\label{sec_experimental_setup}


In our classification task (drug subtype), we utilize 
classifiers such as SVM, Naive Bayes (NB), Multi-Layer Perceptron (MLP), K Nearest Neighbors (KNN), Random Forest (RF), Logistic Regression (LR), and Decision Tree (DT). Our evaluation metrics include average accuracy, precision, recall, weighted F1, macro F1, ROC-AUC, and classifier training runtime. We split our data into random training and test sets with a $70-30 \%$ split, and repeat experiments $5$ times.


We employ linear regression, ridge regression, lasso regression, random forest regression, and gradient boosting regression to predict solubility AlogPS. Our evaluation metrics for this task include mean squared error (MSE), mean absolute error (MAE), root mean squared error (RMSE), coefficient of determination ($R^2$), and explained variance score (EVS).

\subsection{Baseline Methods}

\begin{itemize}
    \item \textbf{Morgan Fingerprint}: The circular Morgan fingerprint~\cite{nakajima2021machine} is used to encode the presence of substructures within a molecule. It generates a binary vector that indicates the presence or absence of substructures.
    \item \textbf{MACCS Fingerprint}: The MACCS fingerprint~\cite{durant2002reoptimization,keys2005mdl} is a binary fingerprint that uses predefined substructures based on functional groups and ring systems commonly found in organic molecules. 
    \item \textbf{$k$-mers}: This method is a sequence-based embedding that encodes the frequencies of overlapping sub-sequences of length $k$~\cite{kang2022surrogate} in the SMILES string. 
    \item \textbf{Weighted $k$-mers}: To enhance the $k$-mers-based embedding's quality, we use a weighted version that employs Inverse Document Frequency (IDF) to assign weights to each $k$-mer within the embedding~\cite{ozturk2020exploring}. 
\end{itemize}

\paragraph{Dataset Statistics:}
We used 2 datasets for experimentation. First, a set of $6299$ SMILES strings from the DrugBank dataset~\cite{shamay2018quantitative} is used. To classify the drugs, we assigned drug subtypes (totaling $188$ distinct subcategories) as target labels. For regression analysis, we used solubility AlogPS. The top $10$ drug subcategories, extracted from the Food and Drug Administration (FDA) website~\footnote{\url{https://www.fda.gov/}}, are presented in Table~\ref{tbl_data_stats}. 
The second dataset consists of a set of $16395$ SMILES strings from the ChEMBL~\cite{chemBL_website} dataset where we classify these sequences for $51$ Standard Type and regression task for AlogP value. The top 10 Standard Types and their count, Maximin, and minimum string length are presented in Table~\ref{tbl_data_stats}.

\begin{table}[h!]
    \centering
    \resizebox{0.83\textwidth}{!}{
    \begin{tabular}{@{\extracolsep{4pt}}ccccc || ccccc}
    \toprule
    \multicolumn{5}{c}{Drug Bank Dataset} & \multicolumn{5}{c}{ChEMBL Dataset}\\
    \cmidrule{1-5} \cmidrule{6-10}
    & & \multicolumn{3}{c}{String Length Statistics} & & & \multicolumn{3}{c}{String Length Statistics}\\
    \cmidrule{3-5} \cmidrule{8-10}
     Drug Subcategory & Count & Min. & Max. & Avg. & Standard Type & Count & Min. & Max. & Avg. \\
    \midrule \midrule
         Others & 6299 & 2 & 569 & 55.4448 & IC50 & 4876 & 2 & 248 & 53.3169 \\
 Barbiturate & 54 & 16 & 136 & 51.2407 &  Activity & 2373 & 10 & 169 & 56.7821 \\ 
 Amide Local Anesthetic & 53 & 9 & 149 & 39.1886 &  AC50 & 2201 & 7 & 234 & 50.4371 \\
 Non-Standardized Plant Allergenic Extract & 30 & 10 & 255 & 66.8965 &  RBA & 1421 & 23 & 155 & 53.9078 \\
 Sulfonylurea & 17 & 22 & 148 & 59.7647 &  Ki & 1390 & 2 & 248 & 53.8388 \\
 Corticosteroid & 16 & 57 & 123 & 95.4375 &  EC50 & 1306 & 19 & 114 & 49.1256 \\ 
 Nonsteroidal Anti-inflammatory Drug & 15 & 29 & 169 & 53.6000 &  Potency & 766 & 4 & 248 & 42.6802 \\
 Nucleoside Metabolic Inhibitor & 11 & 16 & 145 & 59.9090 &  Efficacy & 749 & 28 & 107 & 54.2390 \\ 
 Nitroimidazole Antimicrobial & 10 & 27 & 147 & 103.800 &  Inhibition & 456 & 22 & 103 & 53.7478 \\ 
 Muscle Relaxant & 10 & 9 & 82 & 49.8000 &  Emax & 172 & 23 & 96 & 61.5698 \\
        \bottomrule
    \end{tabular}
    }
    \caption{Drug subtypes (top $10$) extracted from FDA website for \textbf{DrugBank dataset} and Standard type (top $10$) extracted from \textbf{ChEMBL dataset}.}
    \label{tbl_data_stats}
\end{table}



\begin{figure}[h!]
  \begin{subfigure}{0.20\textwidth}
  \centering
        \includegraphics[scale=0.045]{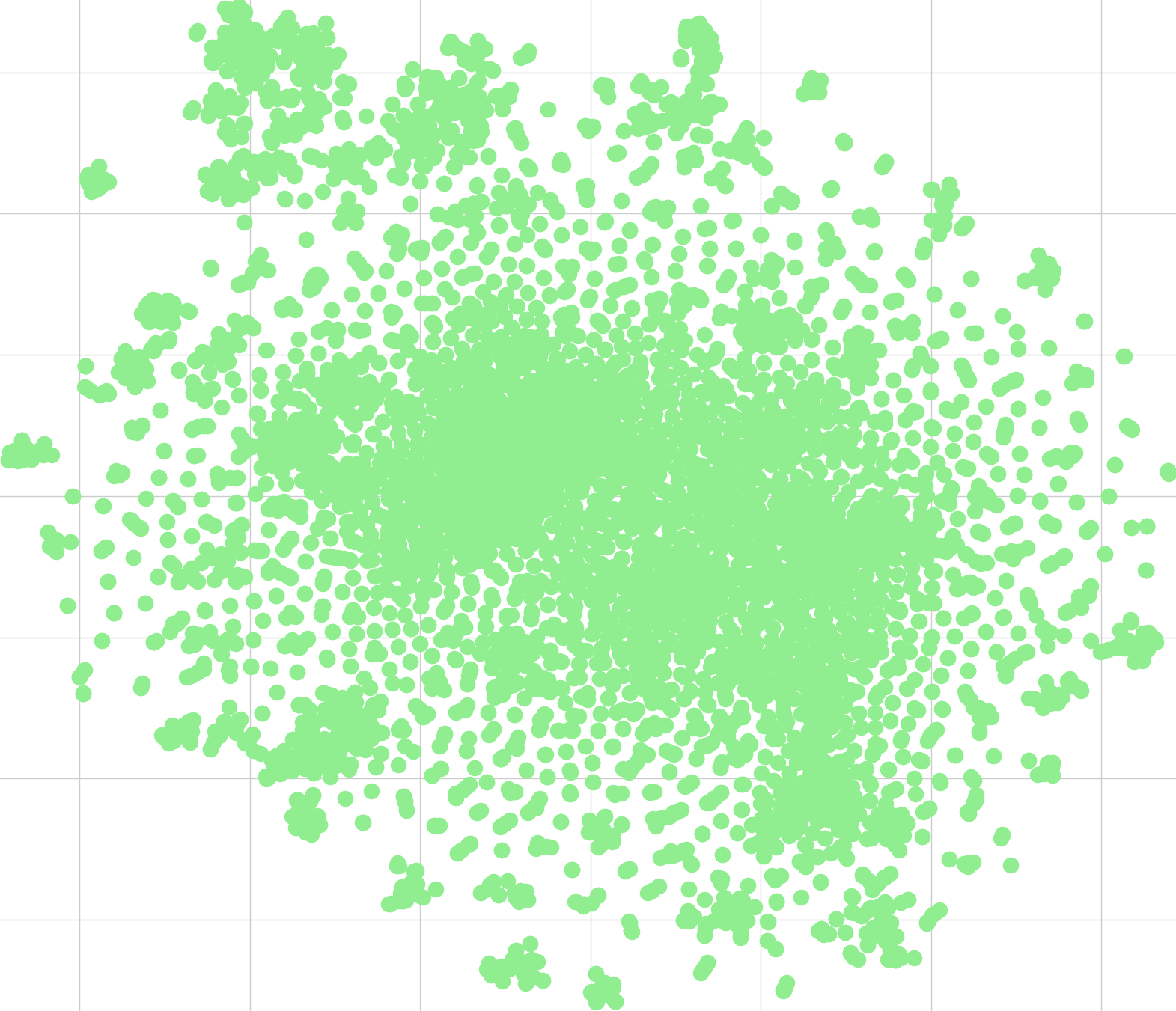}
        \caption{\raggedright Morgan Fingerprint}
        \label{tSNE_Morgan_Fingerprint}%
    \end{subfigure}%
    \begin{subfigure}{0.20\textwidth}
  \centering
        \includegraphics[scale=0.045]{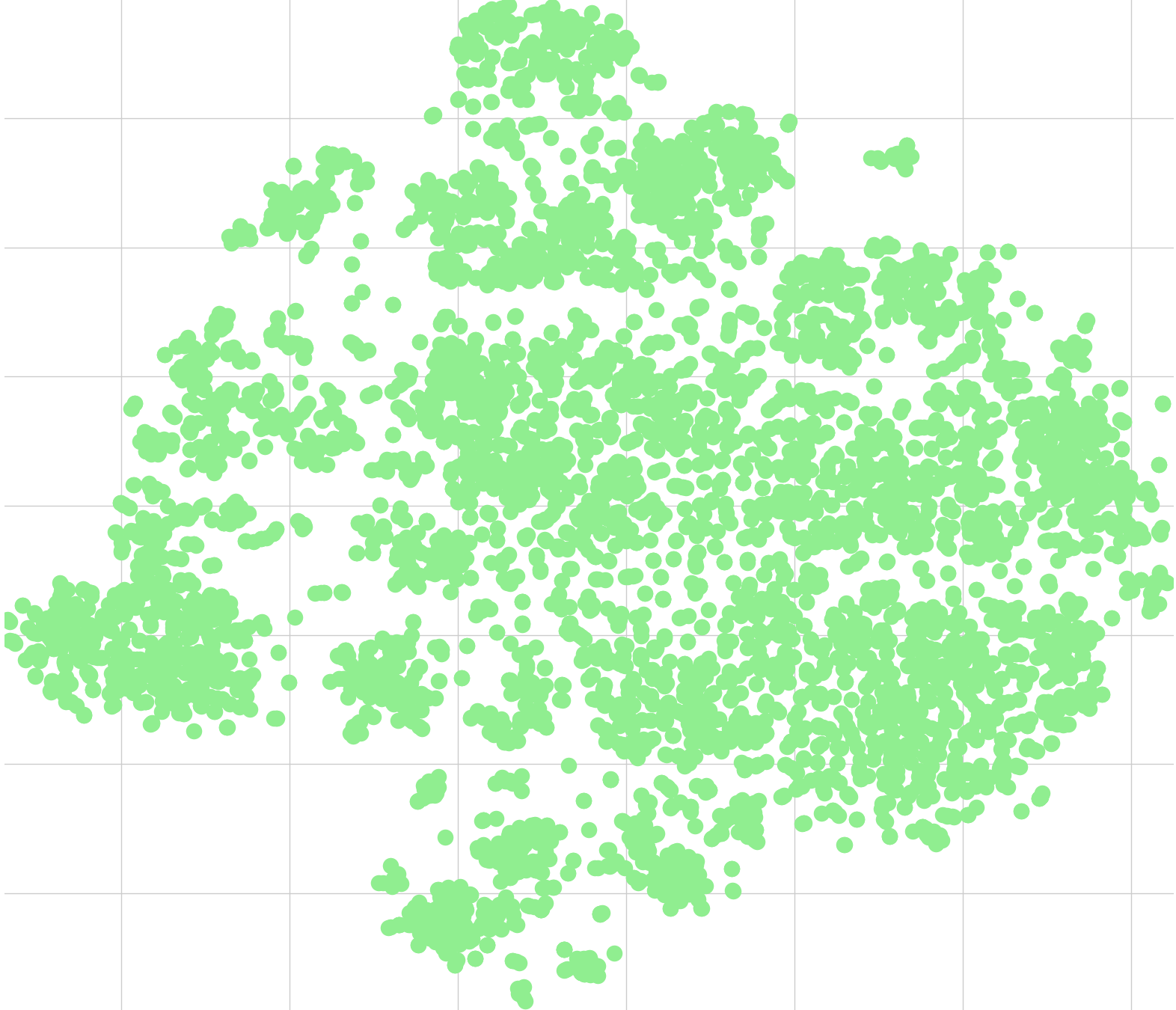}
        \caption{\raggedright MACCS Fingerprint}
        \label{tSNE_MACCS}%
    \end{subfigure}%
    \begin{subfigure}{0.20\textwidth}
  \centering
        \includegraphics[scale=0.045]{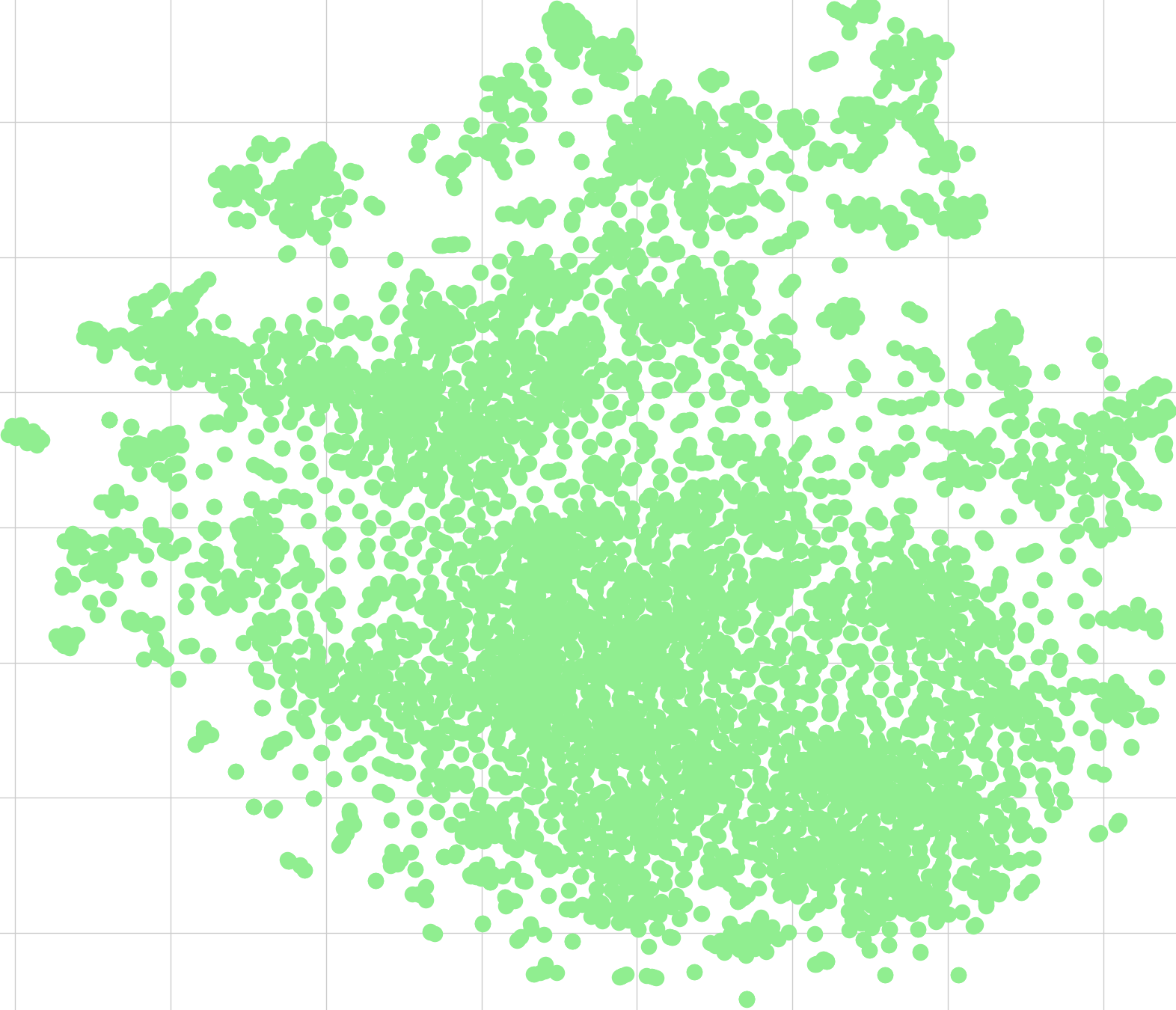}
        \caption{\raggedright $k$-mers Fingerprint}
        \label{tSNE_Spike2Vec}%
    \end{subfigure}%
    \begin{subfigure}{0.20\textwidth}
  \centering
        \includegraphics[scale=0.045]{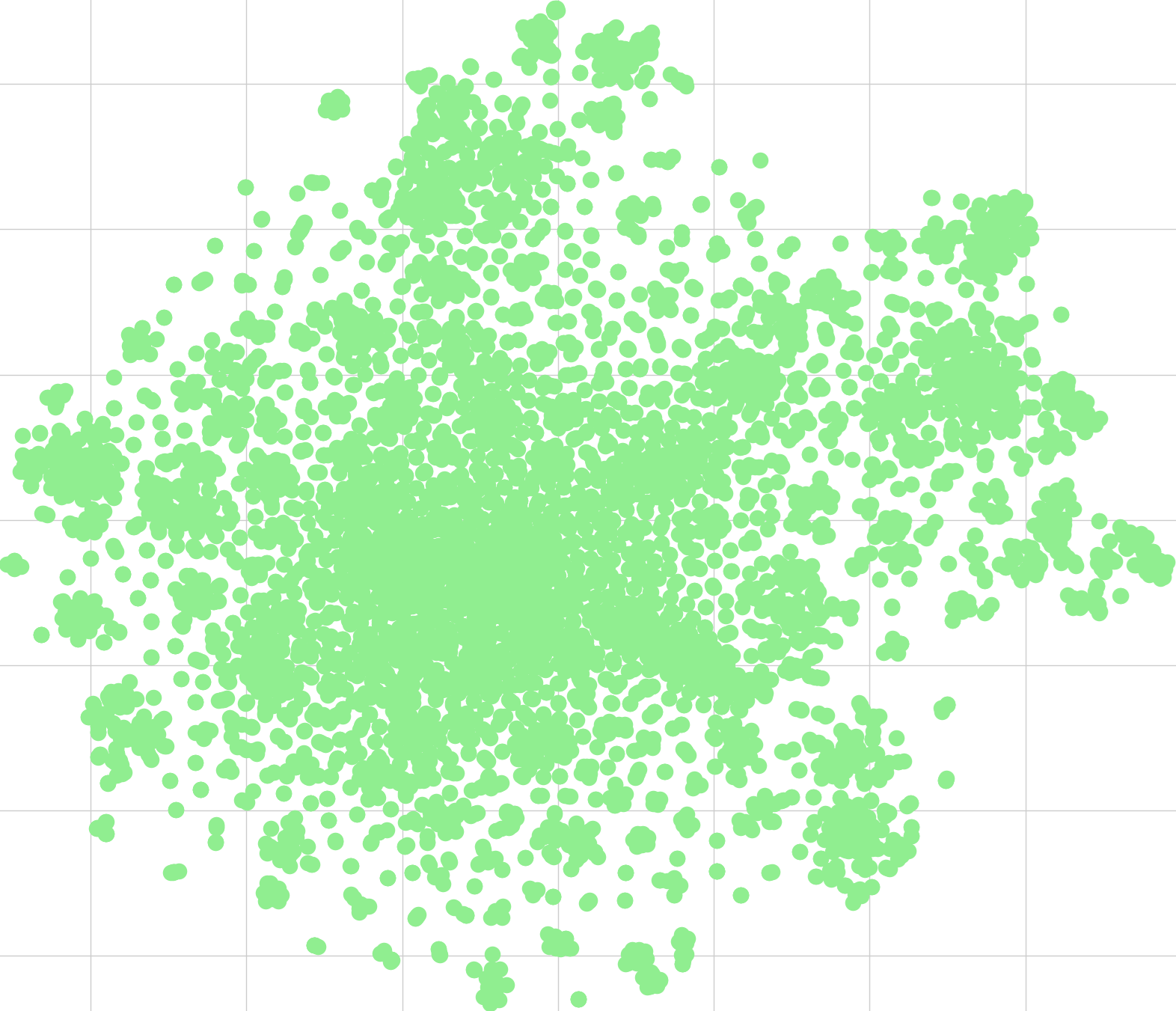}
        \caption{\raggedright Weighted $K$-mers}
        \label{tSNE_Weighted_Kmer}%
    \end{subfigure}%
    \begin{subfigure}{0.20\textwidth}
  \centering
        \includegraphics[scale=0.045]{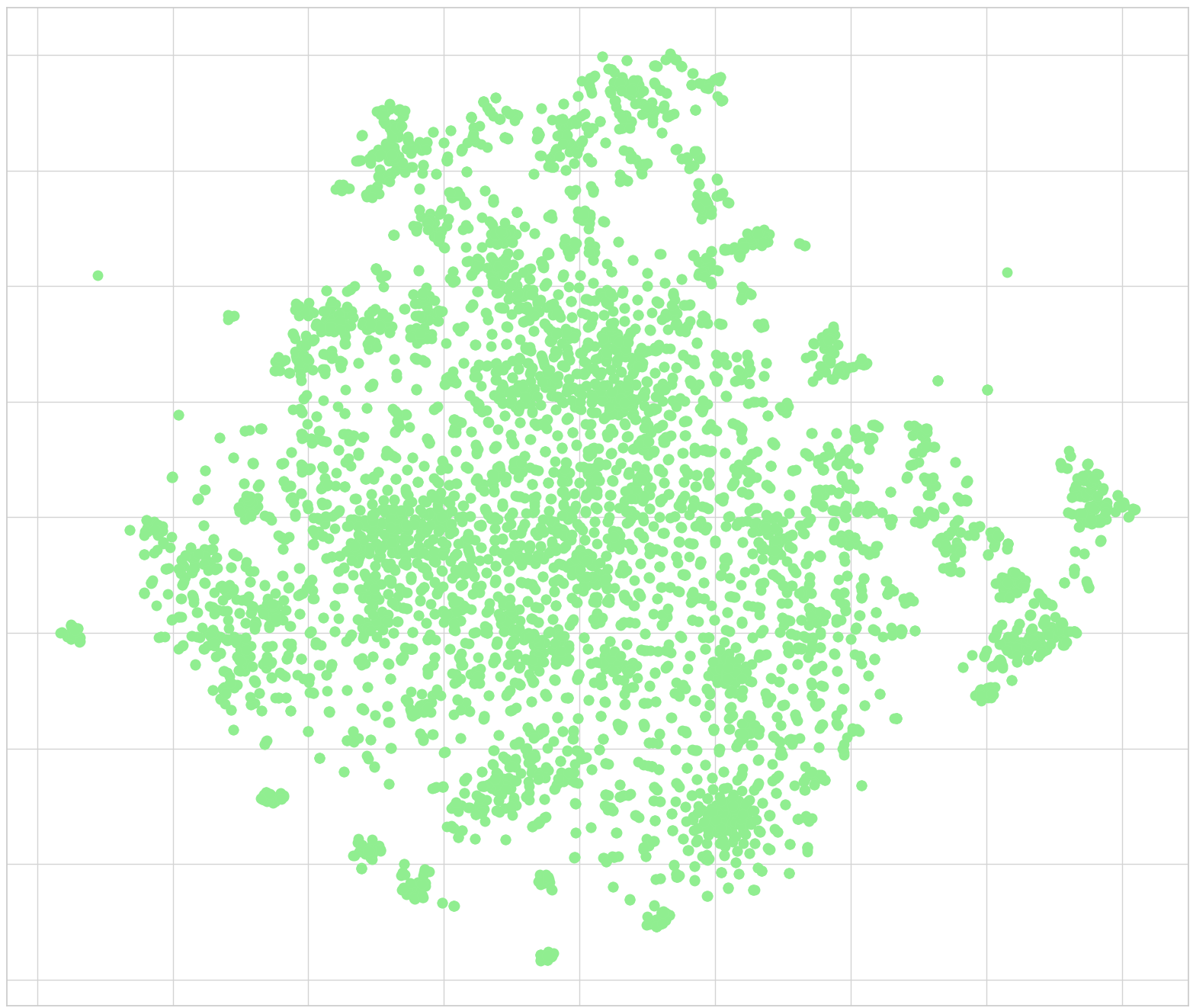}
        \caption{\raggedright SMILES Kernel (ours)}
        \label{tSNE_Smiles_Kernel}%
    \end{subfigure}%
    \caption{The t-SNE plots using feature embedding for the \textbf{Drug Bank} dataset.}
  \label{tsne_plots}
\end{figure}

\paragraph{Data Visualization:}
To assess whether various embedding techniques are maintaining the structure of the data, we employed the t-distributed Stochastic Neighbour Embedding (t-SNE)~\cite{van2008visualizing} algorithm to generate 2-dimensional representations of the embeddings for visual inspection. Figure~\ref{tsne_plots} displays the scatterplots generated by t-SNE for different embedding methods for the drug bank dataset. Overall, the MACCS fingerprint exhibits some clustering, while the proposed SMILES kernel-based approach shows smaller grouping throughout the scatterplot. Similar behavior is observed ChEMBL dataset (we did not include t-SNE plots for the ChEMBL dataset due to page limit constraint).

\section{Results And Discussion}\label{sec_results}
This section reports the classification and regression results for baselines and the proposed kernel approach for the SMILES string analysis.

\paragraph{Classification Results:}
Table~\ref{tble_results_classification} displays the classification results for the DrugBank dataset, indicating that our proposed kernel-based approach outperforms other embedding methods and classifiers in terms of average accuracy, precision, recall, weighted F1 score, and ROC-AUC. The weighted $k$-mers method performs better than other methods in terms of Macro F1. These results indicate that our kernel approach, by projecting data into a higher-dimensional space, enhances the ability of underlying classifiers to differentiate between various types of SMILES strings.
Table~\ref{tble_results_classification} displays the classification results for the ChEMBL dataset. Although Morgan Fingerprint outperforms for accuracy, precision, recall, and weighted F1 score. But we can see for ROC-AUC our proposed method performs better.

\begin{table}[h!]
    \centering
    \resizebox{0.99\textwidth}{!}{
    \begin{tabular}{p{1.8cm}p{1.1cm}p{1.2cm}p{1.2cm}p{1.2cm}p{1.2cm}p{1.3cm}p{1.2cm}p{1.3cm} |
    p{1.2cm}p{1.2cm}p{1.2cm}p{1.2cm}p{1.2cm}p{1.2cm}p{1.2cm}
    }
    \toprule
    & & \multicolumn{7}{c}{Drug Bank Dataset} & \multicolumn{7}{c}{ChEMBL Dataset}\\
    \cmidrule{2-9} \cmidrule{10-16}  
        Embed. & Algo. & Acc. $\uparrow$ & Prec. $\uparrow$ & Recall $\uparrow$ & F1 (Wt.) $\uparrow$ & F1 (Mac.)$\uparrow$ & ROC-AUC $\uparrow$ & Train Time (Sec.) $\downarrow$  & Acc. $\uparrow$ & Prec. $\uparrow$ & Recall $\uparrow$ & F1 (Wt.) $\uparrow$ & F1 (Mac.)$\uparrow$ & ROC-AUC $\uparrow$ & Train Time (Sec.) $\downarrow$ \\
        \midrule \midrule
\multirow{7}{2.1cm}{Morgan Fingerprint}
& SVM & 0.8838 & \underline{0.8577} & 0.8838 & 0.8696 & \underline{0.0591} & \underline{0.5383} & 17.6993 & 0.4427 & 0.4367 & 0.4427 & 0.4369 & 0.2474 & 0.6174 & 216.976 \\
& NB & 0.8969 & 0.8454 & 0.8969 & 0.8697 & 0.0275 & 0.5068 & 3.5027 & 0.2794 & 0.4307 & 0.2794 & 0.2679 & 0.2086 & \underline{0.6765} & 10.2868 \\
& MLP & 0.8297 & 0.8493 & 0.8297 & 0.8390 & 0.0245 & 0.5239 & 17.4977  & 0.4300 & 0.4271 & 0.4300 & 0.4259 & 0.1800 & 0.5834 & 76.8557 \\
& KNN & 0.9129 & 0.8543 & 0.9129 & \underline{0.8795} & 0.0374 & 0.5130 & \underline{0.2560}  & 0.4829 & 0.4812 & 0.4829 & 0.4748 & 0.2606 & 0.6229 & \underline{4.1277} \\
& RF & 0.9109 & 0.8499 & 0.9109 & 0.8764 & 0.0258 & 0.5088 & 3.4253 & 0.4764 & 0.4686 & 0.4764 & 0.4695 & 0.2460 & 0.6103 & 55.3410 \\
& LR & \underline{\textbf{0.4934}} & \underline{\textbf{0.4868}} & \underline{\textbf{0.4934}} & \underline{\textbf{0.4870}} & \underline{0.2689} & 0.6215 & 10.2591 & \underline{\textbf{0.4934}} & \underline{\textbf{0.4868}} & \underline{\textbf{0.4934}} & \underline{\textbf{0.4870}} & \underline{0.2689} & 0.6215 & 10.2591 \\
& DT & 0.8569 & 0.8512 & 0.8569 & 0.8534 & 0.0333 & 0.5286 & 1.2680 & 0.4321 & 0.4298 & 0.4321 & 0.4251 & 0.2263 & 0.6074 & 4.3639 \\

\cmidrule{2-9} \cmidrule{10-16}                                                           
\multirow{7}{2.1cm}{MACCS Fingerprint}                                
& SVM & 0.8705 & \underline{0.8539} & 0.8705 & 0.8613 & \underline{0.0520} & \underline{0.5441} & 3.1812 & 0.4676 & 0.4626 & 0.4676 & 0.4418 & 0.2594 & 0.6368 & 76.7914 \\
& NB & 0.2458 & 0.8473 & 0.2458 & 0.3698 & 0.0359 & 0.5224 & 0.5048 & 0.0842 & 0.3224 & 0.0842 & 0.0915 & 0.1396 & \underline{0.7046} & 0.7156 \\
& MLP & 0.8659 & 0.8444 & 0.8659 & 0.8547 & 0.0220 & 0.5175 & 21.0636 & 0.4638 & 0.4415 & 0.4638 & 0.4399 & 0.1893 & 0.5949 & 25.2533 \\
& KNN & 0.9076 & 0.8447 & 0.9076 & 0.8741 & 0.0305 & 0.5107 & \underline{0.0903} & 0.4816 & \underline{0.4771} & 0.4816 & 0.4711 & 0.2352 & 0.6191 & \underline{\textbf{0.3670}} \\
& RF & 0.9057 & 0.8499 & 0.9057 & \underline{0.8749} & 0.0344 & 0.5149 & 1.1254  & \underline{0.4819} & 0.4721 & \underline{0.4819} & \underline{0.4751} & \underline{\textbf{0.2724}} & 0.6380 & 6.6008 \\
& LR & \underline{0.9126} & 0.8331 & \underline{0.9126} & 0.8710 & 0.0100 & 0.5000 & 3.2345 & 0.4426 & 0.4328 & 0.4426 & 0.4186 & 0.2251 & 0.6016 & 8.9684 \\
& DT & 0.8227 & 0.8522 & 0.8227 & 0.8363 & 0.0457 & 0.5436 & \textbf{0.1100} & 0.4444 & 0.4408 & 0.4444 & 0.4368 & 0.2308 & 0.6254 & 0.2999 \\
 \cmidrule{2-9} \cmidrule{10-16}                                                                             
 \multirow{7}{2.1cm}{$k$-mers}                           
 & SVM & 0.8190 & \underline{0.8514} & 0.8190 & 0.8341 & 0.0413 & \underline{0.5487} & 11640.03 & 0.4271 & 0.4164 & 0.4271 & 0.4154 & \underline{0.2265} & 0.6235 & 764.380 \\
 & NB & 0.7325 & 0.8425 & 0.7325 & 0.7816 & 0.0247 & 0.5149 & 2348.88 & 0.0799 & 0.2705 & 0.0799 & 0.0696 & 0.1095 & \underline{0.6881} & \underline{0.9319} \\
 & MLP & 0.8397 & 0.8465 & 0.8397 & 0.8426 & 0.0270 & 0.5311 & 7092.26 & 0.3821 & 0.3603 & 0.3821 & 0.3651 & 0.1176 & 0.5545 & 78.6732 \\
 & KNN & \underline{0.9101} & 0.8480 & \underline{0.9101} & \underline{0.8766} & 0.0429 & 0.5167 & \underline{68.50} & 0.4334 & 0.4284 & 0.4334 & 0.4216 & 0.1657 & 0.5913 & 1.6151 \\
 & RF & \underline{0.4461} & \underline{0.4362} & \underline{0.4461} & \underline{0.4351} & 0.1938 & 0.5874 & 14.1722 & \underline{0.4819} & 0.4721 & \underline{0.4819} & \underline{0.4751} & \underline{\textbf{0.2724}} & 0.6380 & 6.6008 \\
 & LR & 0.8885 & 0.8423 & 0.8885 & 0.8642 & \underline{0.0461} & 0.5286 & 1995.11 & 0.4211 & 0.4057 & 0.4211 & 0.4065 & 0.2015 & 0.5962 & 87.3623 \\
 & DT & 0.8429 & 0.8490 & 0.8429 & 0.8455 & 0.0397 & 0.5361 & 211.38 & 0.3883 & 0.3837 & 0.3883 & 0.3813 & 0.1690 & 0.5809 & 1.5972 \\

\cmidrule{2-9} \cmidrule{10-16}
\multirow{7}{2.1cm}{Weighted $k$-mers}
 & SVM & 0.8219 & 0.8355 & 0.8219 & 0.8368 & 0.0451 & \underline{0.5490} & 9926.76 & 0.4698 & 0.4629 & 0.4698 & 0.4604 & \underline{0.2658} & 0.6409 & 405.782 \\
 & NB & 0.7490 & 0.8475 & 0.7490 & 0.7931 & 0.0360 & 0.5221 & 2564.96 & 0.1929 & 0.2973 & 0.1929 & 0.2008 & 0.1940 & \underline{0.6925} & \underline{1.3470} \\
 & MLP & 0.8288 & 0.8511 & 0.8288 & 0.8392 & 0.0270 & 0.5345 & 7306.79 & 0.4397 & 0.4224 & 0.4397 & 0.4259 & 0.1587 & 0.5835 & 84.402 \\
 & KNN & 0.9122 & 0.8473 & 0.9122 & 0.8728 & 0.0307 & 0.5091 & \underline{53.06} & 0.4626 & 0.4619 & 0.4626 & 0.4536 & 0.2096 & 0.6098 & 1.597 \\
 & RF & \underline{0.9135} & 0.8455 & \underline{0.9135} & \underline{0.8758} & 0.0245 & 0.5067 & 619.65  & 0.4340 & 0.4313 & 0.4340 & 0.4250 & 0.1987 & 0.5867 & 29.458 \\
 & LR & 0.8928 & 0.8492 & 0.8928 & 0.8697 & \underline{\textbf{0.0595}} & 0.5293 & 1788.37 & \underline{0.4786} & \underline{0.4670}
 & \underline{0.4786} & \underline{0.4674} & 0.2566 & 0.6260 & 103.276 \\
 & DT & 0.8420 & \underline{0.8518} & 0.8420 & 0.8461 & 0.0445 & 0.5347 & 147.47 & 0.3666 & 0.3648 & 0.3666 & 0.3605 & 0.1541 & 0.5738 & 8.838 \\
\cmidrule{2-9} \cmidrule{10-16} 
\multirow{7}{2.1cm}{SMILES Kernel}
 & SVM & 0.8430 & 0.8554 & 0.843 & 0.8478 & 0.0519 & 0.5375 & 32.3892 & 0.4436 & 0.4405 & 0.4436 & 0.4398 & 0.2474 & 0.6257 & 535.8723 \\
& NB & 0.6256 & \underline{\textbf{0.8624}} & 0.6256 & 0.7209 & \underline{0.0755} & 0.5412 & 4.1092 & 0.2246 & 0.3327 & 0.2246 & 0.2303 & 0.2290 & \underline{\textbf{0.7053}} & 4.8780 \\
& MLP & 0.8222 & 0.8437 & 0.8222 & 0.8326 & 0.0204 & 0.5078 & 34.097 & 0.4241 & 0.4178 & 0.4241 & 0.4188 & 0.1616 & 0.5751 & 54.1374 \\
& KNN & 0.9116 & 0.8501 & 0.9116 & 0.8783 & 0.0442 & 0.5147 & \underline{0.5929} & \underline{0.4899} & \underline{0.4861} & \underline{0.4899} & \underline{0.4808} & 0.2324 & 0.6104 & \underline{1.8517} \\
& RF & 0.9145 & 0.8580 & 0.9145 & \underline{\textbf{0.8801}} & 0.0324 & 0.5118 & 91.06 & 0.4671 & 0.4597 & 0.4671 & 0.4597 & \underline{0.2331} & 0.6067 & 103.7672 \\
& LR & \underline{\textbf{0.915}} & 0.8372 & \underline{\textbf{0.915}} & 0.8744 & 0.0112 & 0.5001 & 56.4993 & 0.2971 & 0.0882 & 0.2971 & 0.1361 & 0.0108 & 0.5000 & 31.9301 \\
& DT & 0.828 & 0.8499 & 0.828 & 0.8381 & 0.043 & \underline{\textbf{0.5733}} & 59.5459 & 0.4110 & 0.4099 & 0.4110 & 0.4048 & 0.1985 & 0.5968 & 22.0352 \\

         \bottomrule
         \end{tabular}
    }
    \caption{Classification results (of $5$ runs) for different methods using different evaluation metrics on \textbf{DrugBank dataset}. The best values are shown in bold.}
    \label{tble_results_classification}
\end{table}

\paragraph{Regression Results:}
Table~\ref{tbl_regression_results} presents the regression results for the DrugBank dataset, which indicate that the random forest regression model with MACCS fingerprint outperforms all other embedding methods and regression models. Although our proposed kernel-based approach did not perform comparitively, it is still able to achieve results comparable to those obtained using the MACCS fingerprint in combination with random forest regression.
Table~\ref{tbl_regression_ChEMBL_results} presents the regression results for the ChEMBL dataset. We can see Linear regression and Ridge regression models with weighted $k$-mer for our proposed embeddings.

\begin{table}[h!]
    \centering
    \resizebox{0.8\textwidth}{!}{
    \begin{tabular}{@{\extracolsep{4pt}}ccccccc}
    \toprule
        Embedding & Algo. & MAE $\downarrow$ & MSE $\downarrow$ & RMSE $\downarrow$ & $R^2$ $\uparrow$ & EVS $\uparrow$ \\
        \midrule \midrule
\multirow{5}{2.1cm}{Morgan Fingerprint}
& Linear Regression & 63.2345 & 11601.2046 & 107.7088 & 0.3139 & 0.3143 \\
& Ridge Regression & 62.6110 & 11529.2733 & 107.3744 & 0.3182 & 0.3185 \\
& Lasso Regression & 53.4116 & 11043.7095 & 105.0890 & 0.3469 & 0.3474 \\
& Random Forest Regression & 24.0881 & 7722.9372 & 87.8802 & 0.5433 & 0.5439 \\
& Gradient Boosting Regression & 32.4982 &  8853.8418 & 94.0948 & 0.4764 &  0.4768 \\
\cmidrule{2-7}
\multirow{5}{2.1cm}{MACCS Fingerprint}
& Linear Regression & 55.7719 & 11202.9967 & 105.8442 & 0.3375 & 0.3378 \\
& Ridge Regression & 55.5289 & 11167.1285 & 105.6746 & 0.3396 & 0.3399 \\
& Lasso Regression & 54.1349 & 11189.4825 & 105.7803 & 0.3383 & 0.3385 \\
& Random Forest Regression & \textbf{17.8092} & \textbf{3711.9790} & \textbf{60.9260} & \textbf{0.7804} & \textbf{0.7809} \\
& Gradient Boosting Regression & 31.4769 & 7308.5600 & 85.4901 & 0.5678 & 0.5678 \\
\cmidrule{2-7}
\multirow{5}{2.1cm}{$k$-mers}
& Linear Regression & 8.3616e+10 & 4.6111e+23 & 6.7905e+11 & -2.72674e+19 & -2.72670e+19  \\
& Ridge Regression & 59.1402 & 12955.0398 & 113.8202 & 0.2339 & 0.2339  \\
& Lasso Regression & 51.7842 & 12608.1103 & 112.2858 & 0.2544 & 0.2545  \\
& Random Forest Regression & 23.2473 & 6073.5836 & 77.9331 & 0.6408 & 0.6420  \\
& Gradient Boosting Regression & 32.3582 & 8709.4397 & 93.3243 & 0.4849 & 0.4855 \\
\cmidrule{2-7}
\multirow{5}{2.1cm}{Weighted $k$-mers}
& Linear Regression & 1.3608e+11 & 1.6509e+24 & 1.2848e+12 & -9.7624e+19 & -9.7527e+19  \\
& Ridge Regression & 62.8535 & 13187.9852 & 114.8389 & 0.2201 & 0.2202  \\
& Lasso Regression & 55.5155 & 12241.4725 & 110.6411 & 0.2761 & 0.2762  \\
& Random Forest Regression & 24.0294 & 6224.7174 & 78.8968 & 0.6319 & 0.6330  \\
& Gradient Boosting Regression & 33.0856 & 9066.1662 & 95.2164 & 0.4638 & 0.4644 \\
\cmidrule{2-7}
\multirow{5}{2.1cm}{SMILES kernel}
 & Linear Regression & 55.4431 & 10084.3079 & 100.4206 & 0.40368 & 0.40418 \\
 & Ridge Regression & 50.5369 & 16914.8699 & 130.0571 & -0.00023 & 0.000006 \\
 & Lasso Regression & 50.5372 &
16914.9748 &
130.0575 &
-0.00023 &
0.0 \\
 & Random Forest Regression & 23.0957
 & 5056.7441
& 71.1107
& 0.7009
& 0.7023 \\
 & Gradient Boosting Regression & 25.4830
& 5642.1382
& 75.1141
& 0.6663
& 0.6664 \\
         \bottomrule
         \end{tabular}
    }
    \caption{Regression results for different models and evaluation metrics on \textbf{DrugBank dataset}. The best values are shown in bold.}
    \label{tbl_regression_results}
\end{table}

\begin{table}[h!]
    \centering
    \resizebox{0.8\textwidth}{!}{
    \begin{tabular}{@{\extracolsep{4pt}}ccccccc}
    \toprule
        Embedding & Algo. & MAE $\downarrow$ & MSE $\downarrow$ & RMSE $\downarrow$ & $R^2$ $\uparrow$ & EVS $\uparrow$ \\
        \midrule \midrule
\multirow{5}{1.8cm}{Morgan Fingerprint}
& Linear Regression & 0.4782 & 0.5745 & 0.7580 & 0.8793 & 0.8794	\\
& Ridge Regression & 0.4701 & 0.5583 & 0.7472 & 0.8827 & 0.8828	\\
& Lasso Regression & 1.2661 & 3.0094 & 1.7347 & 0.3679 & 0.3681 \\
& Random Forest Regression & 0.2927 & 0.4280 & 0.6542 & 0.9101 & 0.9103	\\
& Gradient Boosting Regression & 0.7672 & 1.1591 & 1.0766 & 0.7565 & 0.7567	\\
\cmidrule{2-7}
\multirow{5}{1.8cm}{MACCS Fingerprint}
& Linear Regression & 2.9302e+06 & 2.8153e+16 & 1.6779e+08 & -5.9133e+15 & -5.9115e+15 \\
& Ridge Regression & 0.8614 & 1.4766 & 1.2152 & 0.6898 & 0.6901	\\
& Lasso Regression & 1.1829 & 2.6943 & 1.6414 & 0.4341 & 0.4343	\\
& Random Forest Regression & 0.2939 & 0.4034 & 0.6351 & 0.9153 & 0.9153	\\
& Gradient Boosting Regression & 0.7530 & 1.1529 & 1.0737 & 0.7578 & 0.7581	\\
\cmidrule{2-7}
\multirow{5}{1.8cm}{$k$-mers}
& Linear Regression & 0.4294 & 0.3504 & 0.5920 & 0.9264 & 0.9264				\\
& Ridge Regression & 0.4292 & 0.3501 & 0.5917 & 0.9265 & 0.9265                 \\
& Lasso Regression & 0.5544 & 0.5588 & 0.7475 & 0.8826 & 0.8826                 \\
& Random Forest Regression & 0.2133 & 0.2067 & 0.4546 & 0.9566 & 0.9566         \\
& Gradient Boosting Regression & 0.4287 & 0.3730 & 0.6108 & 0.9217 & 0.9217     \\
\cmidrule{2-7}
\multirow{5}{1.8cm}{Weighted $k$-mers}
& Linear Regression & \textbf{0.2862} & 0.1521 & \textbf{0.3899} & \textbf{0.9681} & \textbf{0.9681} \\
& Ridge Regression & \textbf{0.2862} & \textbf{0.1520} & \textbf{0.3899} & \textbf{0.9681} & \textbf{0.9681} \\
& Lasso Regression & 0.7781 & 1.0640 & 1.0315 & 0.7765 & 0.7765                 \\
& Random Forest Regression & 0.3257 & 0.4030 & 0.6348 & 0.9153 & 0.9154         \\
& Gradient Boosting Regression & 0.6523 & 0.8361 & 0.9144 & 0.8244 & 0.8244     \\
\cmidrule{2-7}
\multirow{5}{1.8cm}{SMILES kernel}
& Linear Regression & 0.6138 & 0.9016 & 0.9495 & 0.8106 & 0.8110                \\
& Ridge Regression & 1.6340 & 4.7613 & 2.1820 & -0.0001 & 0.0000                \\
& Lasso Regression & 1.6340 & 4.7613 & 2.1820 & -0.0001 & 0.0000                \\
& Random Forest Regression & 0.3335 & 0.5013 & 0.7080 & 0.8947 & 0.8948         \\
& Gradient Boosting Regression & 1.0104 & 1.9996 & 1.4141 & 0.5800 & 0.5803     \\

         \bottomrule
         \end{tabular}
    }
    \caption{Regression results for \textbf{ChEMBL dataset} for different models and evaluation metrics. The best values are shown in bold.
    }
    \label{tbl_regression_ChEMBL_results}
\end{table}

\begin{figure}[h!]
  \centering
  \begin{subfigure}{0.5\textwidth}
     \centering
     \includegraphics[scale=0.11] {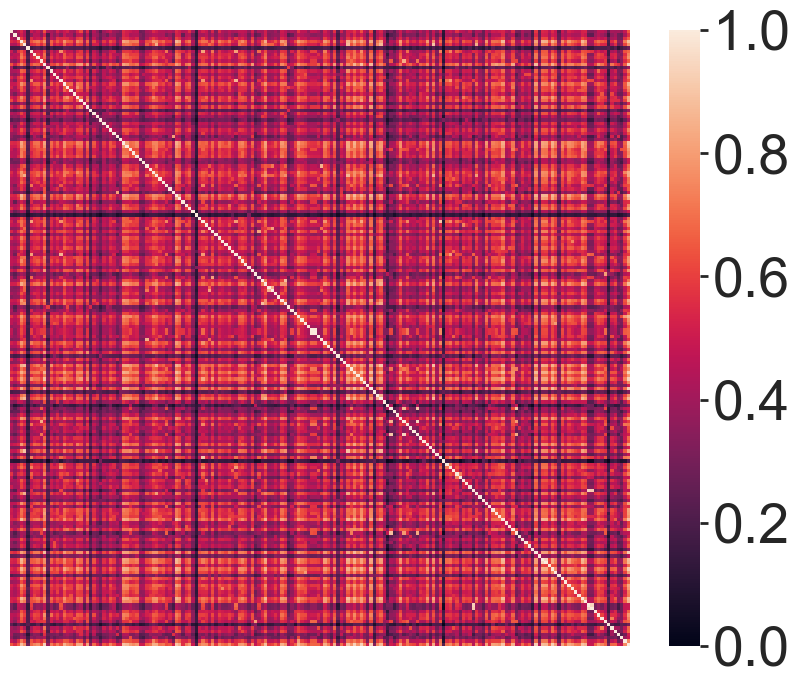}
     \caption{Morgan Fingerprint}
     \label{fig:AAA}
     \end{subfigure}%
     \begin{subfigure}{0.5\textwidth}
  \centering
     \includegraphics[scale=0.11] {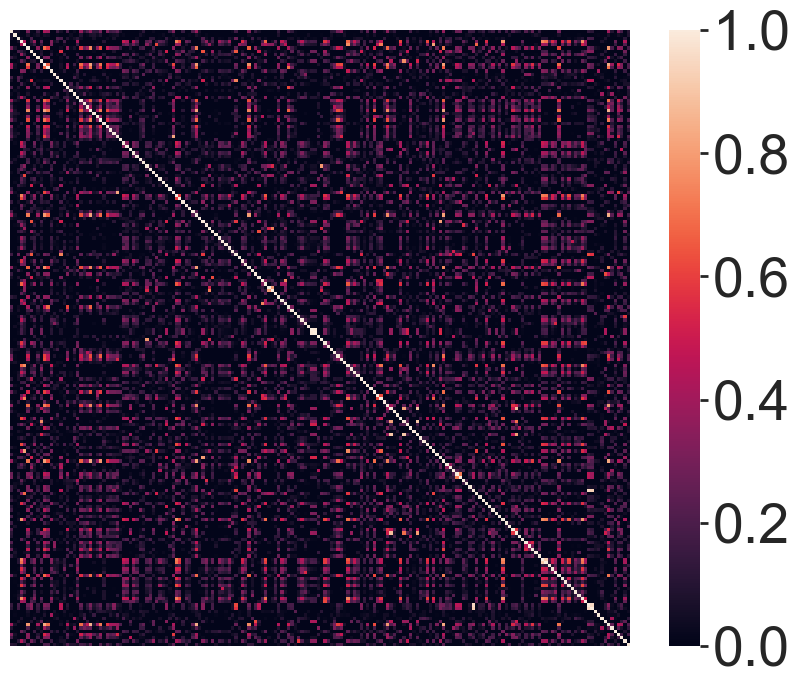}
     \caption{Smiles Kernel}
      \label{fig:BBB}
     \end{subfigure}%
    \caption{Heatmap for classes in \textbf{DrugBank dataset} for different drug subtypes. The figure is best seen in color.}
    \label{fig_heat_map_6897}
\end{figure}

\paragraph{Inter-Class Embedding Interaction:}
We utilize heat maps to analyze further whether our proposed kernel can better identify different classes. These maps are generated by first taking the average of the similarity values to compute a single value for each pair of classes and then computing the pairwise cosine similarity of different class's embeddings with one another. The heat map is further normalized between [0-1] to the identity pattern. The heatmaps for the baseline (i.e., Morgan Fingerprint) and its comparison with the proposed SMILES kernel-based embeddings are reported in Figure~\ref{fig_heat_map_6897}. We can observe that in the case of the Morgan fingerprint, the embeddings for different labels are similar. This eventually means it is difficult to distinguish between different classes due to high pairwise similarities among their vectors. 
On the other hand, we can observe that the pairwise similarity between different class embeddings is distinguishable for proposed SMILES kernel-based embeddings. This essentially means that the embeddings that belong to similar classes are highly similar to each other. In contrast, the embeddings for different classes are very different, indicating that the proposed methodology can accurately identify similar classes and different classes.

\begin{figure}[h!]
  \centering
  \begin{subfigure}{0.4\textwidth}
     \centering
     \includegraphics[scale=0.19] {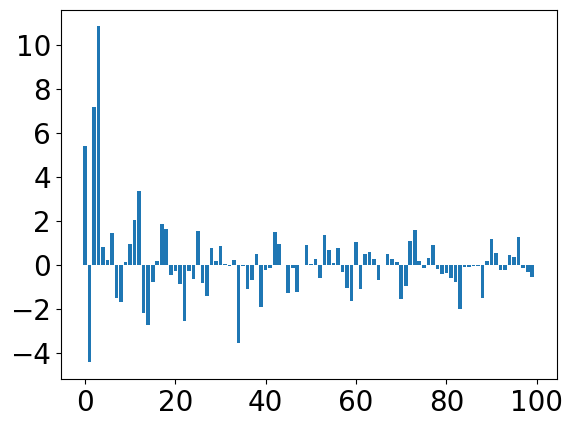}
     \caption{Anti-coagulant}
     \label{fig:AAA}
  \end{subfigure}%
  \begin{subfigure}{0.6\textwidth}
     \centering
     \includegraphics[scale=0.19] {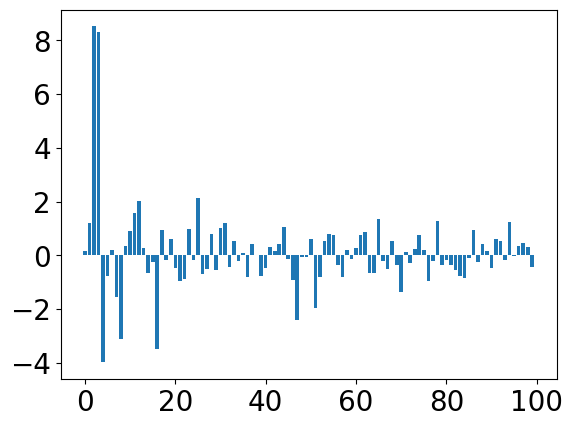}
     \caption{Calcineurin Inhibitor Immunosuppressant}
      \label{fig:BBB}
  \end{subfigure}%
    \caption{Comparing two pairs of classes. (a) and (b) belong to different classes. The Gaussian kernel for (a) and (b) is 0.21 while for the proposed method is 0.17 (a smaller value is better) on \textbf{DrugBank dataset}. Bar plot where we used kernel PCA with k=100 (x-axis) and respective values (y-axis).}
\label{fig_smiles_comparison}
\end{figure}

An example of a pair of sample SMILES strings belonging to different classes (i.e., drug subcategories) is shown in Figure~\ref{fig_smiles_comparison}, where, using morgan fingerprint-based embeddings (a benchmark embedding method from literature) for two random SMILES string samples (i.e., SMILES strings belonging to classes ``Anti-coagulant" and ``Calcineurin Inhibitor Immunosuppressant") as input, we compute kernel value between the embeddings using the typical Gaussian kernel and the proposed SMILES kernel. To get an effective representation, we used kernel PCA and reduced the data dimensionality of the embeddings to $100$ (getting top principal components) before computing the kernel value. Our proposed SMILES kernel, which gave us a smaller value of $0.17$, can capture differences among classes more effectively as compared to the typical Gaussian kernel, which gives us a bigger kernel value of $0.21$ (smaller kernel value is better). The visualizations show sharp spikes, especially in the lower dimensions (closer to 0 on the x-axis). This indicates that there is significant variance in those dimensions, and they are crucial for classification and SMILES kernel can identify and prioritize the variance captured in these initial dimensions better than the Gaussian kernel.  
The later dimensions (closer to 100) have relatively low variance. Better results of our proposed kernel indicate that it can handle noise effectively. The Gaussian kernel assumes a certain shape and structure to the data. It might not always be the best fit for all datasets. Our proposed kernel seems to be more attuned to the characteristics and nuances of the data, leading to better performance.


\paragraph{\textbf{Discussion:}}
Our analysis of solubility and drug subcategory prediction tasks using different embedding methods and models revealed that they perform differently. For drug subcategory prediction, our proposed kernel-based method showed better performance than other embedding methods and classifiers in terms of average accuracy, precision, recall, weighted F1 score, and ROC-AUC. However, weighted $k$-mers outperformed other methods for the Macro F1 score. In contrast, for solubility prediction, the MACCS fingerprint combined with a random forest regression model yielded the best results in terms of multiple evaluation metrics, including RMSE, MAE, and MSE. Although the traditional fingerprint methods performed better for regression analysis, they were not as effective as our proposed approach for classification. Our study provides insights into the effectiveness of different embeddings, which can help researchers choose the most suitable approach for drug discovery applications.


\section{Conclusion}\label{sec_conclusion}
In this paper, we proposed a kernel-based approach for encoding and analyzing molecular structures represented as SMILES strings. 
We evaluated the proposed approach using the SMILES string dataset for molecular property prediction and activity classification. 
The proposed kernel-based approach represents a promising direction for the analysis and design of molecular structures using kernel methods. Further research can explore the use of other types of kernels and the application of the proposed approach to other areas of chemistry and material science. 
We believe that our proposed approach will contribute to the development of new drugs and materials with desirable properties, leading to significant advancements in healthcare and technology. 

\bibliography{references}
\bibliographystyle{splncs04}

\end{document}